# Neural Network Model for Path-Planning Of Robotic Rover Systems

Youssef Bassil

LACSC – Lebanese Association for Computational Sciences
Registered under No. 957, 2011, Beirut, Lebanon
youssef.bassil@lacsc.org

## ABSTRACT

Today, robotics is an auspicious and fast-growing branch of technology that involves the manufacturing, design, and maintenance of robot machines that can operate in an autonomous fashion and can be used in a wide variety of applications including space exploration, weaponry, household, and transportation. More particularly, in space applications, a common type of robots has been of widespread use in the recent years. It is called planetary rover which is a robot vehicle that moves across the surface of a planet and conducts detailed geological studies pertaining to the properties of the landing cosmic environment. However, rovers are always impeded by obstacles along the traveling path which can destabilize the rover's body and prevent it from reaching its goal destination. This paper proposes an ANN model that allows rover systems to carry out autonomous path-planning to successfully navigate through challenging planetary terrains and follow their goal location while avoiding dangerous obstacles. The proposed ANN is a multilayer network made out of three layers: an input, a hidden, and an output layer. The network is trained in offline mode using back-propagation supervised learning algorithm. A software-simulated rover was experimented and it revealed that it was able to follow the safest trajectory despite existing obstacles. As future work, the proposed ANN is to be parallelized so as to speed-up the execution time of the training process.

## 1. INTRODUCTION

Robotics technology is emerging at a rapid pace, offering new possibilities for automating tasks in many challenging applications, especially in space explorations, military operations, underwater missions, domestic services, and medical procedures. Particularly, in space exploration, robotic devices are formally known as planetary rovers or simply rovers and they are aimed at conducting physical analysis of planetary terrains and astronomical bodies, and collecting data about air pressure, climate, temperature, wind, and other atmospheric phenomena surrounding the landing sites [1]. Basically, rovers can be autonomous capable of operating with little or no assistance from ground control or they can be remotely controlled from earth ground stations called RCC short for Remote Collaboration Center [2].

In essence, the movement of autonomous rovers is not directed by human operators; instead, it is controlled by complex algorithms that allow the rover to traverse paths on multiple terrains while avoiding obstacles and path errors. This capability is more formally known as path-planning in which a rover or any robotic vehicle can perform terrain analysis and select the safest route to travel across [3]. The rover can then proceed towards the goal location over the selected trajectory while avoiding obstacles without previous knowledge of their existence.

This paper proposes a path-planning solution for autonomous robotic planetary rover systems based on artificial neural network (ANN) [4]. The proposed neural network is multi-layer consisting of three consecutive layers: an input, a hidden, and an output layer. The input layer is made out of two neurons that are fed by the rover's sensors which are designed to detect obstacles of any size and shape. The hidden layer is made out of three neurons and its purpose is to read input data and multiply them by a certain weight and then forward the results to the next layer. The output layer is made out of two neurons that are directly linked to the rover's motors which control its movement and its mechanical operation. The proposed ANN uses a mix of activation functions including Sigmoid for the hidden neurons and linear for the output neurons. Moreover, the model employs a supervised learning approach using the back-propagation algorithm [5] to train the network in offline mode.

The proposed artificial neural network is meant to allow the rover system selects the best path through any given ground by predicting the existing obstacles along the path and the harsh structure of the landing terrain. This would allow the rover to navigate autonomously and safely toward its goal location and complete its designed task.

## 2. SPACE EXPLORATION ROVERS

Fundamentally, a rover is a space exploration robotic vehicle used particularly in exploring the land of a planet. It has the capability to travel across the surface of a landscape and other cosmic bodies. A rover has many features: It can generate power from solar panels; capture high-resolution images; move in 360 degrees with the help of a navigation camera (Navcam); walk across obstacles such as bumps and rocks; conduct deep analysis and record measurements using multiple types of spectrometers; find properties of materials to identify their types and their composition; search for geological clues such as water to detect any presence of life on the landing environment; and inspect the mineralogy and texture of the local terrain using panoramic cameras (Pancam) [6][7]. Financially, robotic rovers can cost to build, test, and deploy hundreds of millions of dollars sometimes billions of dollars [8]. Historically, Lunokhod and Marsokhod were two space rovers designed and launched by the soviets in the 70s [9]; while, Spirit and Opportunity were two US rovers produced by NASA, the space agency of the United States, between year 2004 and 2010 as part of NASA's ongoing Mars Exploration Rover Mission (MER).

Inherently, there exist two types of rover vehicles: The first type is the human-controlled rovers which are remotely manipulated from earth and usually guided to perform a particular operation. Communication between the rover and the earth control occurs through the Deep Space Network (DSN), which is an international network of large antennas with communication facilities that supports interplanetary spacecraft missions. Currently, DSN comprises three deep-space communications facilities located in Mojave Desert in California, west of Madrid in Spain, and south of Canberra in Australia.

The second type is the autonomous rovers which can complete their desired tasks without constant human direction. Space exploration rovers are distinguished by a high degree of autonomy as they can cope with their changing environment, automatically gain information about the landing sites, survive a disaster or a failure, operate for prolonged periods of time, execute predefined operations, and navigate across unstructured terrains without human assistance.

In practice, autonomous rovers are most of the time based on artificial evolution to reinforce learning within their environments. This method of learning is known as machine learning in which a system can learn from experience data to generalize so that it performs correctly in new and unseen situations. Predominantly, artificial neural network (ANN), one of the most successfully applied machine learning approaches in robotics field, is used to control autonomous rover systems and provide them an intelligent navigation behavior [10]. In effect, ANN allows the rover to plan and execute collision-free motions within its environment and to reach its goal location while avoiding obstacles and dangerous cosmic objects.

## 3. ARTIFICIAL NEURAL NETWORKS

Artificial neural networks or ANNs for short are very influential brain-inspired computational models, which have been employed in various areas such as computing, medicine, engineering, economics, and many others. ANNs are composed of a certain number of simple computational elements called neurons, organized into a structured graph topology made out of several consecutive layers and immensely interconnected through a series of links called the synaptic weights. Synaptic weights are often associated with variable numerical values that can be adapted so as to allow the ANN to change its behavior based on the problem being tackled [11].

Training an artificial neural network is usually done by feeding the network's input with a pattern to learn. The network then transmits the pattern through its weights and neurons until it generates a final output value. Afterwards, a training or a learning algorithm compares the produced output value to an expected output and if the error range is high, the algorithm marginally alters the network's weights so that if the same pattern is fed again to the network, the output error would be smaller than the previous iteration. This process gets repeated for many cycles called epochs using different set of input patterns until the network produces acceptable outputs for all inputs [12]. This learning progression allows the network to identify many patterns and further generalize to new and unseen patterns. Such type of training is called supervised learning which uses classified pattern information to train the network in offline mode. On the other hand, there exists what so called the unsupervised learning which uses only minimum information without pre-classification to train the network while being in online mode [13]. Some of the most successful supervised learning approaches are feed-forward and back-propagation; while, the most successful unsupervised learning approaches are the Hebbian and the competitive learning rule.

## 4. ADVANTAGES OF ANN IN SPACE APPLICATIONS

Artificial neural networks have many advantages in space applications due to the following reasons [14]:

Generality: Usually, a space rover system is required to process a very high number of parameters that are variable, complex, and received from multiple sources. Neural networks can handle such large-scale problems as it is able to classify objects well even when the distribution of objects in the $N$-dimensional parameter space is very complex.

Performance: Due to the nature of neural networks in executing in a parallel fashion, they can solve problems with multiple constraints and large number of data elements at high-speed and simultaneously. Using parallel technology, rover systems can increase their responsiveness and quickness in detecting, identifying, and handling new patterns behaviors.

Adaptability: Due to the dynamic and always-evolving conditions and challenges in space exploration missions, space rovers are always faced with new trends and patterns. Neural networks can cope with such circumstances as they are adaptable to unseen situations and have the capability to learn data, identify new patterns, and detect trends. A process that is too complex to be achieved by traditional computational techniques

Low energy consumption: Commonly, rovers are powered by solar panels which generate energy from light and photons particles, and then store it into internal batteries with limited lifetime and capacity. A supervised-trained neural network often learns and adjusts its synaptic weights in offline mode; thus, relieving the rover from carrying out insensitive mathematical computations at runtime and consequently reducing processing power, energy, and power consumption.

Robustness & Fault Tolerance: Since in a space applications a cosmic ray can be very destructive, it has however a little impact on neural networks as it can only destroy a few of the neurons, but not the thousands and the millions of neurons which would be able to compensate for the damage, and therefore the output of the network would not be significantly affected.

## 5. RELATED WORK

There are a large variety of solutions already developed for robotic rover systems that are based on artificial neural networks. Several of them are examined in this section:

[15] proposed a neural network model for robust control of space robots. The proposed ANN uses a radial-basis-function (RBF) to handle the various system uncertainties. Besides, the Lyapunov unsupervised learning algorithm is used to train the network and adapt its parameters in online mode. [16] proposed an adaptive approach for controlling robot manipulators using neural networks. The controller is based on the Gaussian radial-basis-function (GRF) which is meant to provide uniformly stable adaptation and asymptotically tracking for the robotic vehicle. The system features a robust controller to overcome the neural network modeling errors and the bounded instabilities. [17] proposed a robust model for space robots based on fuzzy neural network (FNN) controller. The tracking controller can attain very accurate goals in the presence of uncertainties without using linear parameterization and fixed-base robot manipulators. [18] proposed a model for solving the wheel slip problem in space rover exploration devices. The model employs a high fidelity traversability analysis (HFTA) algorithm with path and energy cost functions to predict and detect possible slips while the rover is moving. As a result, the rover can choose the best route via any given topography by avoiding high slip paths. [19] proposed an intelligent model for moving robots with translational and rotational motion deployed in partially structured environment. The model is based on using two neural networks: The first network is utilized to identify the open space using ultrasound range finder data; while, the second network is utilized to identify a safe route for the robot to move across while avoiding the nearby obstacles. [20] presented an advanced robotic architecture for the Rover Mars robot. The system uses advanced infrared sensors coupled with an artificial neural network which is trained using a supervised learning technique. The property of this model is that it uses evolutionary robotics techniques in which an evolvable threshold is used for the rover's sensors. Its purpose is to alter the activation range of infrared sensors in order to differentiate between rocks and holes from the noise originating from landing terrain. The results were a successful rover able of doing several autonomous operations. [21] proposed a neural network approach for robotic systems based on the Jordan architecture. In this approach, the robot can learn at runtime the different patterns of the environment using an internal recurrent artificial neural network. The robot can predict through a series of input sensors the different objects before it. It can then generate navigation steps based on the output signal of the network that would drive the rover's motors devices.

## 6. THE PROPOSED NEURAL NETWORK MODEL

This paper proposes an artificial neural network (ANN) model for robotic planetary rover systems to accomplish path-planning on harsh and bumpy terrains. Its aim is to let the rover vehicle navigates through and follows its intended goal location while avoiding collisions with obstacles, rocks, holes, and sharp slopes of arbitrary shape and size. The movement of the rover is fully autonomous as it is totally controlled by the neural network without any human assistance except when training the network in offline mode.

The proposed ANN is a three layers "2-3-2" network composed of an input, a hidden, and an output layer. The input layer is made out of 2 input source nodes x and y with two corresponding neurons which are physically fed by the rover's sensors. The hidden layer is made out of 3 neurons which receive input data from the input layer and multiply them by the values of the synaptic weights denoted by $W_{ij}$ and then forward the resulted values to the output layer. The output layer is made out of 2 neurons that are directly linked to the rover's motors which control its movement and its mechanical operation.

The employed activation function is Sigmoid for the hidden neurons; whereas, it is linear for the output neurons. The synaptic weights range from $W_{00}$ to $W_{13}$ and they represent the interconnection between the different neurons of the network. Additionally, two biases are employed in the hidden and the output layers to regulate and limit the output of the network and they are denoted by $b_h$ and $b_o$.

Formally, the proposed neural network can be defined as follows:

$NN = \{ I, T, W, A \}$ where I denotes the set of input nodes, T denotes the topology of the network including the number of layers and the number of their neurons, W denotes the set of synaptic weights values, and A denotes the activation function.

$I = \{x, y\}$
$T = \{ L_{in-2}, L_{h0-3}, L_{out-2} \}$
$W_{Lin} = \{ W_{00}, W_{01}, W_{02}, W_{10}, W_{11}, W_{12}, W_{20}, W_{21}, W_{22} \}$
$W_{h0-3} = \{ W_{00}, W_{01}, W_{02}, W_{03}, W_{10}, W_{11}, W_{12}, W_{13} \}$
$A = \{1/1+e^{-t}, 1\}$

Figure 1 illustrates the architecture of the proposed ANN.

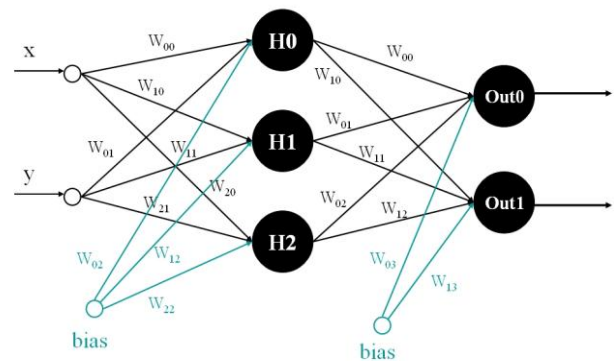

**Figure 1:** ANN architecture

### 6.1 The Back-Propagation Algorithm

The proposed ANN model is trained through a supervised learning approach using the back-propagation algorithm [5]. The back-propagation algorithm comprises two passes: A forward pass which propagates the input data in the forward direction from the input layer to output layer of the network. The pass eventually ends up by generating an output value and computing an error value while leaving synaptic weights intact. In effect, the error is calculated by subtracting the desired output from the actual output just generated. If the error is within an acceptable range, then the network is trained with new set of input data; otherwise, a backward pass is executed. The backward pass is a reverse pass which propagates the error signal backward through the network layers so as to update the synaptic weights of the network. The different steps of the back-propagation algorithm can be summarized as follows:

1. Feed the network with an input vector and a corresponding desired output vector.
2. Calculate the output of the network using forward pass.
3. Calculate the output error signal.
4. If error is within an acceptable range, move to the next input vector, otherwise go backward and update the weights of the network.
5. Keep repeating the above steps until all input vectors are consumed

Additionally, and in order to attain more accurate results, the back-propagation algorithm was fine-tuned with extra parameters whose purpose is to regulate and add more accuracy to the learning process by shifting the activation function to the left or to the right. The controlling parameters are listed below:

- Biases: $b_h$ and $b_o$
- Learning Parameter: $\eta$
- Momentum Alpha Parameter: $\alpha$

### 6.2 Computing the Back-Propagation Algorithm

Computationally, the proposed model is governed by the following steps and mathematical equations:

1. Perform the forward propagation and calculate the output signal using *(X1 * W1)+(X2 * W2)+...+(Xi * Wi)+...+ (Xn * Wn)*
2. Calculate the Sigmoid activation function *(1 / 1+e^{−input})* for the hidden neurons and the linear activation function *(y = v)* for the output neurons.
3. Calculate the error using *error = desired output – actual output*
4. Perform the back propagation algorithm using the following equations to update the weights of the network:

   Case1: For the output neurons:
   *Weight (n+1) = Weight (n) + α[ΔWeight(n-1)] + (N * Output (previous neuron) * error)*

   Case2: For hidden neurons:
   *Weight (n+1) = Weight (n) + α[ΔWeight(n-1)] + [N * Output (previous neuron) * Output (this current hidden neuron) * (1-Output (this current hidden neuron)) * $\sum_k$ $error_k$ * $weight_{kj}$ ]*

### 6.3 The Learning Process

Although the back-propagation algorithm looks simple, computing it is quite an intensive task as it requires a series of arithmetic operations executed for hundreds of iterations. Below are the various calculations required to train the proposed network using the back-propagation algorithm.

Initial Weights:
$W_{00}$ = 0.17 ; $W_{01}$ = 0.33 ; $W_{02}$ = 0.1 ; $W_{10}$ = 0.3 ; $W_{11}$ = 0.71 ; $W_{12}$ = 0.21 ; $W_{20}$ = 0.15 ; $W_{21}$ = 0.43 ; $W_{22}$ = 0.69
$W_{00}$ = 0.11 ; $W_{01}$ = 0.03 ; $W_{02}$ = 0.52 ; $W_{03}$ = 0.41 ; $W_{10}$ = 0.93 ; $W_{11}$ = 0.14 ; $W_{12}$ = 0.79 ; $W_{13}$ = 0.66

Input Vector: [0 , 0]
Desired Output Vector: [1 , 1]
Learning parameter: 0.25
Biases: +1

Forward Pass:

Input of $h_0$: (0*0.17) + (0*0.33) + (1*0.1) = 0.1
Output of $h_0$: 1/ 1+$exp^{-0.1}$ = 0.524

Input of $h_1$: (0*0.3) + (0*0.71) + (1*0.21) = 0.21
Output of $h_1$: 1/ 1+$exp^{-0.21}$ = 0.552

Input of $h_2$: (0*0.15) + (0*0.43) + (1*0.69) = 0.69
Output of $h_2$: 1/ 1+$exp^{-0.69}$ = 0.665

Input of $O_0$: (output of $h_0$ * 0.11) + (output of $h_1$ * 0.03) + (output of $h_2$ * 0.52) + (1 * 0.41) = (0.524*0.11) + (0.552*0.03) + (0.665*0.52) + (1*0.41) = 0.05764 + 0.01656 + 0.3458 + 0.41 = 0.83
Output of $O_0$: 0.83 *1 = 0.83 (Linear activation Function)

Input of $O_1$:  = (0.524*0.93) + (0.552*0.14) + (0.665*0.79) + (1*0.66) = 0.48732 + 0.07728 + 0.52535 + 0.66 = 1.74995
Output of $O_1$:  = 1.74995 * 1 = 1.74995 (Linear activation Function)

Calculating Error for $O_0$: desired – actual = 1 - 0.83 = 0.17
Calculating Error for $O_1$: desired – actual = 1 - 1.74994 = 0. 74994

Back Propagation:

Starting with output Neuron → Case 1 in the algorithm

For $W_{00}$: weight (new) = weight (old) + ( η * output(previous neuron) * error)= 0.11 + (0.25 * output ($h_0$) * error ($O_0$)) = 0.11 + (0.25 * 0.524*0.17) = 0.13227
For $W_{10}$: 0.93 + (0.25 * 0.524 * error ($O_1$))= 0.93 + (0.25 * 0.524 * (-0.74994)) = 0.83176

For $W_{01}$: 0.03 + (0.25 * output ($h_1$) * error ($O_0$)) = 0.03 + (0.25 * 0.552 * 0.17) = 0.05346

…..etc

Now dealing with the hidden Neurons → Case 2 in the algorithm

For $W_{00}$:
Weight (new) = weight (old) + [η *output (previous neuron) * output (this neuron) * (1 - output(this neuron)) * $\Sigma_k$ error$_k$ * weght$_{kj}$]
Weight (n+1) = 0.17 + [0.25 * input x * output($h_0$) * (1-output($h_0$)) * (error ($O_0$) * $W_{10}$ + error ($O_1$) * $W_{10}$)]
= 0.17 + [0.25 * 0 * 0.524 * (1- 0.524) *((0.17*0.13227) + (-0.74994 * 0.83176))] = 0.17 + 0 = 0.17 = $W_{00}$(n+1)

…..etc

## 7. IMPLEMENTATION

The proposed neural network model was implemented using MS C#.NET 2008 with over 600 lines of code. It was compiled under the MS Visual Studio 2008 and the MS .NET Framework 3.5. It encompasses a training engine that can be fed with various static and dynamic parameters including initial weights, input data, learning rate, momentum, and the number of epochs to execute. Figure 2 depicts the GUI interface of the training engine while executing the back-propagation algorithm to train the neural network.

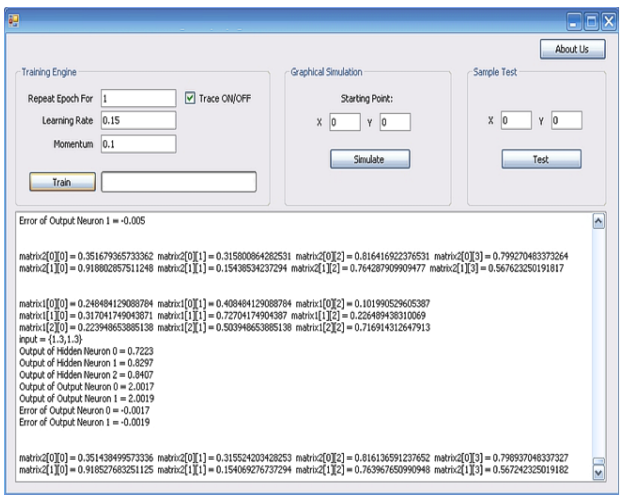

**Figure 2:** Training Engine for the Proposed ANN

Furthermore, and in order to validate the proposed neural network, the rover vehicle was simulated using a software model able to plot trajectories, plan for a certain path, and move towards its goal location.

A sample test case was tested to verify if the rover is able to move from a given point A to a given point B while avoiding obstacles. A random point A was selected as the rover's initial position and is denoted by A[x=0 ; y=0]. Another point B was selected as the rover's goal location and is denoted by B[x=11.73 ; y=0]. Additionally, an obstacle was set along the x-axis preventing the rover from heading directly toward its destination goal in a straight line. The test case was executed and it clearly revealed that the rover was able to reach its goal location without passing through the introduced obstacle. In fact, the rover followed two levels of routes. One ascending that started from [x=0.8287 ; y=0.8287] and ended at [x=5.672 ; y=5.898], and one descending that started from [x=5.672 ; y=5.898] and ended at B[x=11.73 ; y=0], the actual goal location. Figure 3 graphically plots the passage taken by the rover from its initial position A to its goal location B.

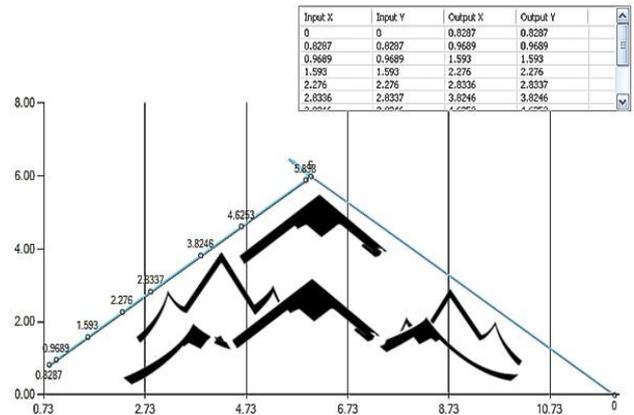

**Figure 3:** Simulation for Path-Planning

Due to the large code base behind the training engine and the rover simulation software, only the implementation of the function responsible for performing the backward pass of the back-propagation algorithm is listed below:

```
private void BackPropagation()
{
    // Case 1: BACKPROPAGATION for OUTPUT NEURONS
    double learningRate = Convert.ToDouble(learningRateTextbox.Text.Trim());
    double momentum = Convert.ToDouble(momentumTextbox.Text.Trim());
    for (int i = 0; i < matrix2.GetLength(0); i++)
    {
        for (int j = 0; j < matrix2.GetLength(1); j++)
        {
            if (j < 3) // j<3 --> 3 hidden neurons
            {
                // Calculating ΔW(n)
                double deltaWeight = (momentum * delta2[i, j]) + (learningRate *
                        outputOfHiddenNeurons[j] * error[i]);

                // Calculating W(n+1)
                matrix2[i, j] = matrix2[i, j] + deltaWeight;

                // Updating the ΔW(n) matrix
                delta2[i, j] = deltaWeight;
            }
            else
            {
                // Calculating ΔW(n)
                double deltaWeight = (momentum * delta2[i, j]) + (learningRate * bias *
                        error[i]);

                // Calculating W(n+1)
                matrix2[i, j] = matrix2[i, j] + deltaWeight;

                // Updating the ΔW(n) matrix
                delta2[i, j] = deltaWeight;
            }
            if (traceON == true)
                outputTextbox.AppendText("matrix2[" + i + "][" + j + "] = " + matrix2[i, j] + "  ");
        }
        if (traceON == true)
            outputTextbox.AppendText("\r\n");
    }
    if (traceON == true)
        outputTextbox.AppendText("\r\n\r\n");
```

```
    // Case 2: BACKPROPAGATION for HIDDEN NEURONS

    for (int i = 0; i < matrix1.GetLength(0); i++)
    {
        for (int j = 0; j < matrix1.GetLength(1); j++)
        {
            // Calculating ΔW(n)
            double deltaWeight = (momentum * delta1[i, j]) + (learningRate * input[index, j] *
            outputOfHiddenNeurons[i] * (1 - outputOfHiddenNeurons[i]) * (error[0] *
                                          matrix2[0, i] + error[1] * matrix2[1, i]));

            // Calculating W(n+1)
            matrix1[i, j] = matrix1[i, j] + deltaWeight;

            // Updating the ΔW(n) matrix
            delta1[i, j] = deltaWeight;

            if (traceON == true)
                outputTextbox.AppendText("matrix1[" + i + "][" + j + "] = " + matrix1[i, j] + "  ");
        }

        if (traceON == true)
            outputTextbox.AppendText("\r\n");
    }
}
```

## 8. CONCLUSIONS & FUTURE WORK

This paper presented an artificial neural network model for robotic rover systems to perform autonomous path-planning during space exploration missions. The proposed ANN is a three-layer network composed of three layers: an input, a hidden, and an output layer. The network is trained through a supervised learning approach using the back-propagation algorithm. The purpose of the model is to control the movement of space rovers allowing them to travel across planetary surfaces while avoiding obstacles in a complete autonomous manner. Experiments conducted showed that a software-simulated rover was able to avoid collision with obstacles and reached its goal location through the safe and correct trajectory.

As future work, the back-propagation algorithm, used in training the proposed network, is to be parallelized so as to take advantage of parallel and distributed computing platforms and speed-up the execution time of the training process.

## ACKNOWLEDGMENTS

This research was funded by the Lebanese Association for Computational Sciences (LACSC), Beirut, Lebanon, under the "Service Oriented Architecture Robotics Research Project – SOARRP2012".

## REFERENCES

[1] Amel Zerigui, Xiang Wu, Zong-Quan Deng, "A Survey of Rover Control Systems", *International Journal of Computer Sciences and Engineering Systems*, Vol. 1, No. 4, pp. 105-109, 2007.

[2] Roland Siegwart, Illah R. Nourbakhsh, Davide Scaramuzza, *Introduction to Autonomous Mobile Robots*, 2nd ed, The MIT Press, 2011.

[3] Jean-Claude Latombe, *Robot Motion Planning*, Kluwer Academic Publishers, 1991.

[4] Chowdhury, F.N., Wahi, P., Raina, R., Kaminedi, S., "A survey of neural networks applications in automatic control", *Proceedings of the 33rd Southeastern Symposium on System Theory*, 2001.

[5] Paul J. Werbos, *Beyond Regression: New Tools for Prediction and Analysis in the Behavioral Sciences*, PhD thesis, Harvard University, 1974.

[6] Mars Exploration Rover, NASA Facts, National Aeronautics and Space Administration, Jet Propulsion Laboratory, California Institute of Technology Pasadena, 2004.

[7] Sarah Loff, NASA's Space Exploration Vehicle (SEV), NASA Official: Rocky Lind, 2011.

[8] Amy Svitak, "Cost of NASA's Next Mars Rover Hits Nearly $2.5 Billion.", 2011, http://www.space.com/10762-nasa-mars-rover-overbudget.html

[9] Wesley T. Huntress JR., Mikhail Ya Marov, *Soviet Robots in the Solar System: Mission Technologies and Discoveries*, Springer, 2011.

[10] Jun Wang, Xiaofeng Liao, Zhang Yi, "Advances in Neural Networks", *Second International Symposium on Neural Networks*, Chongqing, China, 2005.

[11] Simon Haykin, *Neural Networks: A Comprehensive Foundation*, Prentice Hall, 2 ed, 1998.

[12] Simon Haykin, *Neural Networks and Learning Machines*, 3rd ed., Prentice Hall, 2008.

[13] Sotiris B. Kotsiantis, "Supervised Machine Learning: A Review of Classification Techniques", *Informatica*, Vol. 31, No. 3, pp. 249-268, 2007.

[14] Martin Anthony, Peter L. Bartlett, *Neural Network Learning: Theoretical Foundations*, Cambridge University Press, 2009.

[15] Baomin Feng, Guangcheng Ma, Weinan Xie, Changhong Wang, "Robust tracking control of space robot via neural network", *1st International Symposium on Systems and Control in Aerospace and Astronautics*, 2006.

[16] Shuzhi S. Ge Hang, C.C. Woon, L.C., "Adaptive neural network control of robot manipulators in task space", *IEEE Transactions on Industrial Electronics*, 1997.

[17] Changhong Wang, Baomin Feng, Guangcheng Ma, Chuang Ma, "Robust tracking control of space robots using fuzzy neural network", *IEEE International Symposium on Computational Intelligence in Robotics and Automation*, 2005.

[18] Livianu, Mathew Joseph, "Human-in-the-loop neural network control of a planetary rover on harsh terrain", Thesis, Georgia Institute of Technology, 2008.

[19] Danica Janglová, "Neural Networks in Mobile Robot Motion", *International Journal of Advanced Robotic Systems*, Vol. 1, No 1, pp. 15-22, 2004.

[20] Martin Peniak, Davide Marocco, Angelo Cangelosi, "Autonomous Robot Exploration Of Unknown Terrain: A Preliminary Model Of Mars Rover Robot", *In Proceedings of 10th ESA Workshop on Advanced Space Technologies for Robotics and Automation*, Noordwijk, The Netherlands, 2008.

[21] Tani.J, "Model-based Learning for Mobile Robot Navigation from the Dynamical Systems Perspective", *IEEE Trans. on Syst., Man and Cyb.*, Vol. 26, No.3, pp. 421-436, 1996.